\documentclass{article}

\usepackage[nonatbib,preprint]{neurips_2019}

\usepackage[utf8]{inputenc} 
\usepackage[T1]{fontenc}    
\usepackage{hyperref}       
\usepackage{url}            
\usepackage{booktabs}       
\usepackage{amsfonts}       
\usepackage{nicefrac}       
\usepackage{microtype}      

\usepackage{color}
\usepackage{amsmath}
\usepackage{amsthm}
\usepackage{amsfonts}
\usepackage{multirow}
\usepackage{multicol}
\usepackage{amssymb}
\usepackage{enumerate}
\usepackage{xspace}
\usepackage{euscript}
\usepackage{graphicx,epstopdf}
\usepackage{subfig}
\usepackage{amscd}
\usepackage{longtable}
\usepackage{algorithm}
\usepackage{algorithmic}

\newcommand{\argmin}{\mathrm{argmin}}

\newcommand{\st}{\mathrm{~s.t.~}}
\newcommand{\vones}{\mathbf{1}}

\newcommand{\Tr}{\mathrm{Tr}}

\newcommand{\R}{\mathbb{R}}

\newcommand{\cH}{\mathcal{H}}

\newcommand{\cP}{\mathcal{P}}

\newcommand{\cR}{\mathcal{R}}

\newcommand{\cU}{\mathcal{U}}
\newcommand{\cN}{\mathcal{N}}

\newcommand{\dist}{\mathrm{dist}}

\newcommand{\hU}{\hat{U}}

\newtheorem{Theorem}{Theorem}[section]

\newtheorem{Proposition}{Proposition}[section]
\newtheorem{Definition}{Definition}[section]

\usepackage{abstract}

\usepackage{tabularx}

\title{Big-Data Clustering:
	K-Means or K-Indicators?}

\author{%
	Feiyu Chen\thanks{School of Big Data and Software Engineering, Chongqing University, Chongqing, China} \\
	\texttt{fchen@cqu.edu.cn} \\
	\And
	Yuchen Yang\thanks{Department of Computational and Applied Mathematics, Rice University, Houston, Texas, U.S.A.}  \\
	\texttt{yuchen.yang@rice.edu} \\
	\AND
	Liwei Xu\thanks{School of Mathematical Sciences, University of Electronic Science and Technology of China, Chengdu, Sichuan, China} \\
	\texttt{xul@uestc.edu.cn} \\
	\And
	Taiping Zhang\thanks{College of Computer Science, Chongqing University, Chongqing, China} \\
	\texttt{tpzhang@cqu.edu.cn} \\
	\And
	Yin Zhang\footnotemark[2]~~\thanks{Institute for Data and Decision Analytics, The Chinese University of Hong Kong-Shenzhen, China}\\
	\texttt{yinzhang@cuhk.edu.cn} \\
}

\begin{document}
	
	\maketitle
	
	\begin{abstract}
		The K-means algorithm is arguably the most popular data clustering method, commonly applied to processed datasets in some "feature spaces", as is in spectral clustering. Highly sensitive to initializations, however, K-means encounters a scalability bottleneck with respect to the number of clusters K as this number grows in big data applications. In this work, we promote a closely related model called K-indicators model and construct an efficient, semi-convex-relaxation algorithm that requires no randomized initializations.  We present extensive empirical results to show advantages of the new algorithm when K is large.  In particular, using the new algorithm to start the K-means algorithm, without any replication, can significantly outperform the standard K-means with a large number of currently state-of-the-art random replications.  
	\end{abstract}
	
	\section{Introduction}
	
	Clustering analysis is a fundamental unsupervised machine learning strategy with broad-ranging applications, aiming to group unlabelled data objects into clusters according to a certain similarity measure so that objects within each cluster are more similar to each other than otherwise.
	
	Many clustering algorithms have been investigated in the past decades~\cite{cheng1995mean,rodriguez2014clustering,shah2017robust}.  In practice, clustering-friendly datasets rarely occur in nature, which makes it necessary to employ a two-step strategy. First, the raw data was kernelized~\cite{scholkopf1998nonlinear,dhillon2004kernel} or otherwise preprocessed with dimension reduction methods, such as principal component analysis~\cite{ding2004k}, non-negative matrix factorization~\cite{xu2003document,ding2010convex}, spectral embeddings~\cite{shi2000normalized,ng2002spectral,zha2001spectral}, deep auto-encoders~\cite{peng2016deep,ji2017deep,shaham2018spectralnet} or generative adversarial networks~\cite{chen2016infogan}. Second, a clustering algorithm is applied to the latent embedding. Many clustering methods exist for the doing the second step, including K-means~\cite{macqueen1967some,lloyd1982least}, hierarchical clustering~\cite{sibson1973slink}, affinity propagation~\cite{frey2007clustering} and BIRCH~\cite{zhang1996birch}, etc, among which the classic K-means is arguably the method of choice in general situations.  
	
\begin{figure}
		\begin{center}
			\subfloat[Clustering Accuracy]{\includegraphics[width=0.45\textwidth]{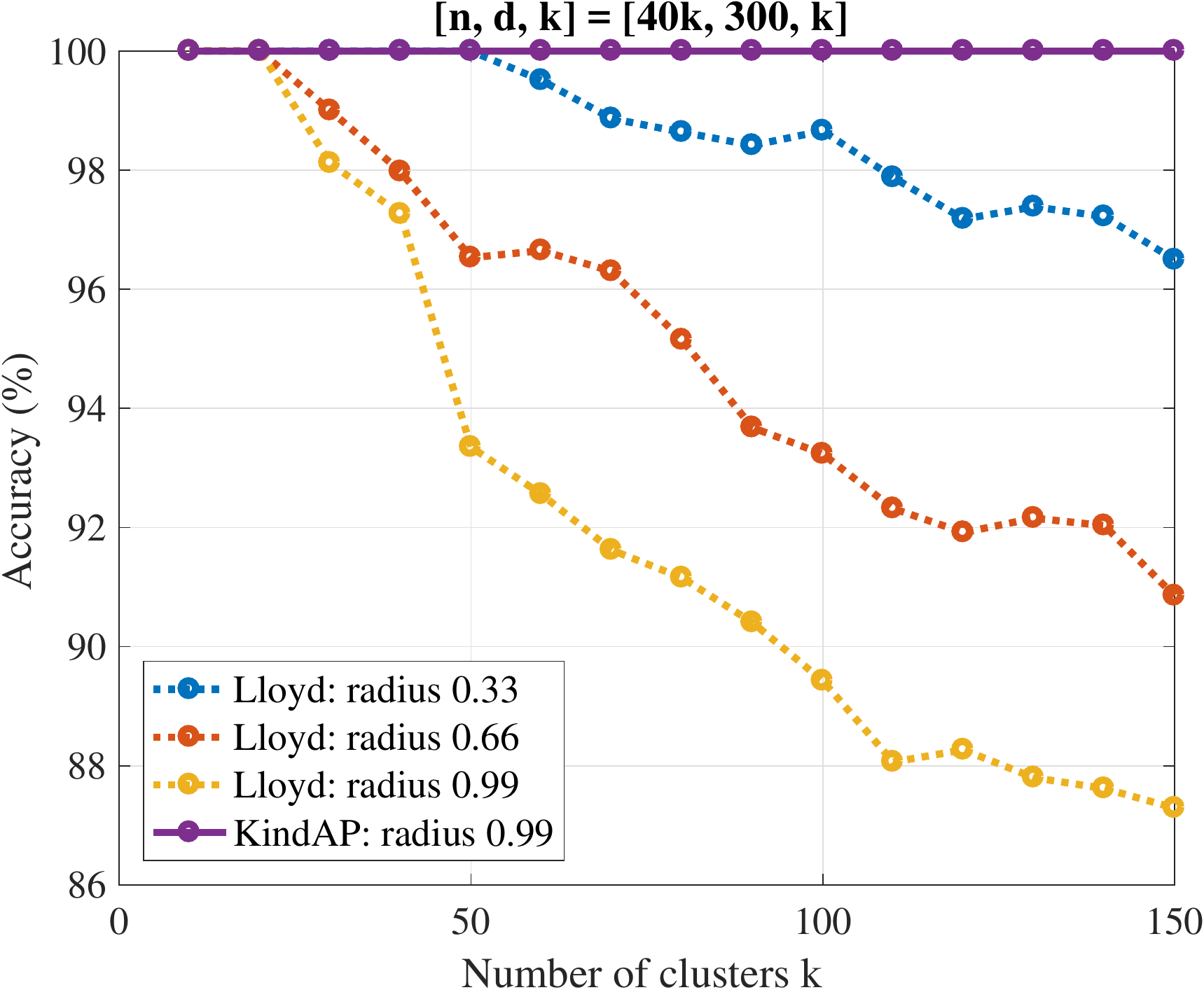}}
			\subfloat[Running Time]{\includegraphics[width=0.45\textwidth]{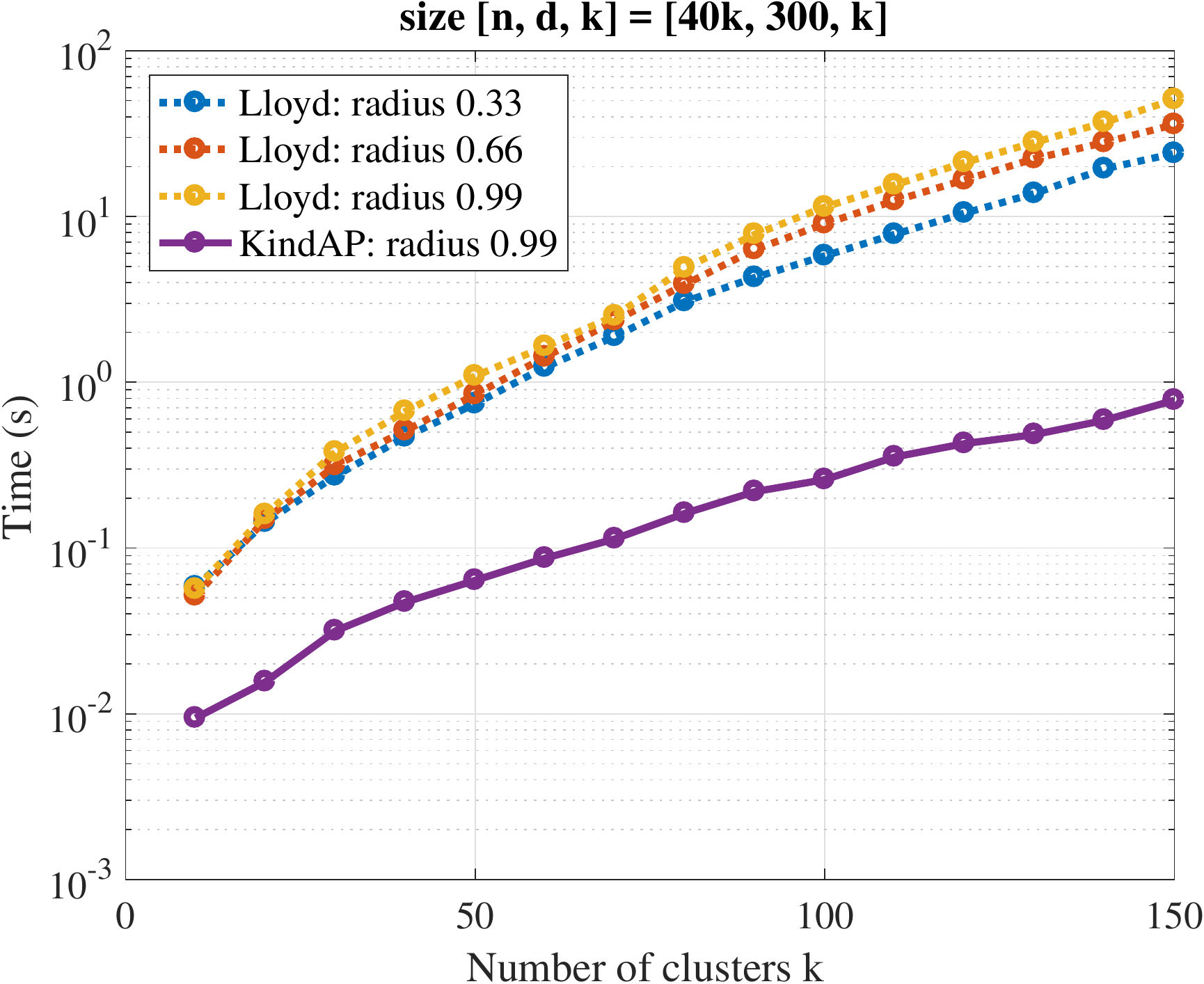}} 
			\vspace{.1cm}
			\caption{
				Synthetic data clouds: we select $k$ center locations in $\R^{300}$, where $k$ varies from $10$ to $150$, such that the distance between each pair of centers is exactly 2. Then $40$ data points are randomly placed on a sphere of radius $\rho = \{0.33,0.66,0.99\}$ around each center to form a cluster. The processed datasets consist of rows of the $n\times k$ matrix formed by the $k$ leading singular vectors of the $n\times 300$ data matrix for $n=40k$. The Lloyd algorithm (with 10 random replications) and the proposed KindAP algorithm are applied to the processed data matrices. Clustering accuracy and running time are recorded.
			}
			\label{fig:side}
		\end{center}
	\end{figure}

	Unfortunately, even with well-processed data the K-means algorithm (also called Lloyd algorithm) still encounters a scalability bottleneck.  It is demonstrated in Figure~\ref{fig:side} that the solution quality of the K-means algorithm deteriorates as the number of clusters increases, while a newly proposed algorithm, called KindAP to be introduced soon, correctly recover the ground truth solutions in all tested cases.  This set of experiments is performed on synthetic datasets with separable clusters (see more details in the caption of Figure~\ref{fig:side}). Later we will show that similar phenomena occur in real image datasets as well.
	
	The root cause of the scalability bottleneck is that greedy algorithms like K-means are highly sensitive to initializations and rely on multiple random replications to achieve good results.  As K increases, the number of random replications needed for good results appears to rise out of control.   To overcome this difficulty, some convex optimization models have been developed. Recent works include semi-definite programming(SDP) ~\cite{peng2007approximating,awasthi2015relax} and linear programming (LP) relaxations \cite{awasthi2015relax} of K-means, and convex fusion methods ~\cite{lindsten2011just,hocking2011clusterpath,chi2015splitting}.  However, the per-iteration complexity of these convex models has been elevated to being quadratic in the number of total samples instead of being linear as in K-means.
	
	In the framework of spectral clustering, an algorithm called spectral rotation (SR) was proposed as an alternative to K-means algorithm~\cite{yu2003multiclass,huang2013spectral} for doing clustering in embedded spaces.    It was argued that the spectral rotation model would be less sensitive to initializations than K-means.   Nevertheless, our experiments (see Section~\ref{section:numerical}) indicate that, at least in some cases, the spectral rotation algorithm could be as sensitive as the K-means.
	
	The main contributions of this paper are summarized as the following:
	\begin{itemize}
		
		\item A general clustering framework is proposed that directly solves for K indicators by subspace matching.   Under suitable conditions, K-means model and the model that we promote are two special cases of the general clustering framework.
		
		\item  A semi-convex-relaxation scheme, called KindAP, is constructed to efficiently solve the particular K-indicators model. KindAP, which is essentially deterministic, can find high-quality solutions at a per-iteration complexity linear in the number of data points.
		
		\item KindAP solutions can be used to warm-start the K-means algorithm without any replication, resulting in better results (measured by the K-means objective) than running K-means with a huge number of K-means++ random replications, when K is relatively large.

		\item Extensive numerical results show a superior scalability of the proposed approach over both K-means and spectral rotation models, especially when the number of clusters becomes large.    

	\end{itemize}
	
	\section{Preliminary}
	This section provides important definitions and concepts for our study. It also revisits the classical K-means model from an unusual angle that will motivate our new K-indicators model. 
	
	\subsection{A set of indicator matrices}
	Consider the problem of clustering a dataset of $n$ objects into $k$ clusters. A matrix $H\in\R^{n\times k}$ is called an indicator matrix if:
	\begin{equation} \label{indicator matrix}
	H_{ij}  =\left\{ 
	\begin{aligned}
	& c_{ij} > 0, \quad \mathrm{~object~} i \in \mathrm{cluster~} j \\
	& 0,   \; \quad \quad \quad \quad \quad \mathrm{~otherwise~}
	\end{aligned}
	\right.
	\end{equation}
	where a positive element $H_{ij} = c_{ij}$ indicates that object $i$ belongs to cluster $j$.   This set of indicator matrices  is the most general, containing  various subsets corresponding to different definitions of indicator matrices in the literature.  For example, $H$ is called a {\em binary indicator matrix} if $c_{ij} \equiv 1$~\cite{yu2003multiclass}, and a {\em normalized indicator matrix} if $c_{ij} \equiv 1/\sqrt{n_j}$, where $n_j$ denotes the number of objects in cluster $j$~\cite{boutsidis2009unsupervised}. 
	
	For convenience, by default we define the set of indicator matrices as:
	\begin{Definition}
		(The set of indicator matrices)
		\begin{equation}\label{indicator set}
		\cH = \left\{H \in \R^{n\times k}:  H^{T}H = I, \; H \ge 0, \; \|e_i^TH\|_0=1 \right\}
		\end{equation} 
	\end{Definition}
	
	Clearly, $\cH$ is a discrete set since each row of $H$ can have only one positive element. 
	To emphasize the pre-determined number $k$, we will refer the $k$ columns of $H$ collectively as K-indicators. These two terms, indicator matrix and K-indicators, will be used exchangeably. 
	
	\subsection{K-means model viewed as subspace-matching}
	
	Although the classic K-means model is commonly written in terms of $k$ centroids, it can also be written in terms of indicator matrices, or K-indicators.  Let $\hU \in \R^{n\times d}$ be a given data matrix where each data vector in $\R^d$ corresponds to a row.  It is well known that the classical K-means model can also be rewritten as~\cite{boutsidis2009unsupervised}: 
	\begin{equation}  \label{Kmeans}
	\min_{H} \quad \|\hU - HH^T\hU \|_F^2  \quad ~\st~   H \in \cH_0 := \cH \cap \left\{H ~|~ \; HH^T\vones_n = \vones_n \right\}
	\end{equation}
	where the constraint $HH^T\vones_n = \vones_n$, together with other constraints in $\cH$, forces non-zero elements in each column of $H$ to have the same value (i.e., $1/\sqrt{n_j}$). In other words, $\cH_0$, the subset of $\cH$, contains all normalized indicator matrices.
	
	
	
	When the data matrix $\hU \in \R^{n \times d} (d \ge k)$ is orthonormal, i.e. $\hU^T \hU = I$, then after some simple calculations the K-means model (\ref{Kmeans}) can be reduced to the following three equivalent optimization problems:
	\begin{equation}\label{K-means3models}
	\min_{H\in\cH_0} \|\hU\hU^T - HH^T\|_F^2 
	~~~\Leftrightarrow~~~ \max_{H\in\cH_0} \| \hU^TH\|_F^2 
	~~~\Leftrightarrow~~~ \max_{H\in\cH_0} \sum_{j=1}^k\sigma_j^2(\hU^TH)
	\end{equation}
	where $\sigma_j(\cdot)$ denotes the $j$-th singular values of a matrix.   We note that taking square root of these objective functions does not change the equivalence.
	
	These relationships provide a subspace-matching perspective for the K-means model.  To see this, we note that the distance of two subspaces can be measured by a distance between their unique orthogonal projections.   For the first model in \eqref{K-means3models}, the two orthogonal projections involved are $\hU\hU^T$ and $HH^T$, respectively.  On the other hand, minimizing a subspace distance under some norms is equivalent to maximizing vector norms of the cosines of principle angles between the two subspaces~\cite{bjoerck1971numerical}.  In \eqref{K-means3models}, these cosines of principle angles are the singular values of $\hU^TH$ where both $\hU$ and $H$ are orthonormal bases.

	\section{K-indicators Model}
	
	The subspace-matching perspective of the K-means model can be extended to a more general framework that solves for an indicator matrix.   We will call it K-indicators framework:
	\begin{equation}  \label{general K-indicators}
	\min_{H} \; \dist(\cR(\hU),\cR(H)), ~\st~   H \in \cH
	\end{equation}
	where $\cR(\hU)$ refers to the range space of $\hU$ (similarly for $\cR(H)$), and "$\dist$" is a subspace distance, which can also be replaced by distance squared, for example.    Clearly, in the K-means model (\ref{Kmeans}) the squared distance function $\|\hU\hU^T - HH^T\|_F^2$ is quartic in $H$ and non-convex.  Can we have a simpler distance function in $H$?
	
	\begin{Theorem}\label{Theorem1}
		If $\hU \in \R^{n \times k}$ such that $\hU^T\hU=I$, and $H \in \R^{n \times k}$ such that $H^TH=I$,  then 
		\begin{equation}\label{dist2}
		dist(\cR(\hU),\cR(H)) := \min_{R^TR = I} \{\|\hU R - H\|_{F} : R \in \R^{k \times k}\}
		\end{equation}
		defines a distance between $\cR(\hU)$ and $\cR(H)$,
		\begin{equation}\label{K-indi3models}
		\min_{H \in \cH} \; \dist^2(\cR(\hU),\cR(H)) ~~\Leftrightarrow~~ \max_{H\in\cH} \| \hU^TH\|_* ~~\Leftrightarrow~~ \max_{H\in\cH} \sum_{j=1}^k\sigma_j(\hU^TH).
		\end{equation} 
		Moreover, 
		\begin{equation}\label{equivalence}
		\frac{\sqrt{2}}{2}\|\hU\hU^T - HH^T\|_F \le \min_{R^TR = I} \|\hU R - H\|_{F}  \le \|\hU\hU^T - HH^T\|_F
		\end{equation}
	\end{Theorem}
	The proof of this theorem is provided in the supplementary material.

	
	Now, we propose the following model based on the distance function (\ref{dist2}):
	\begin{equation}  \label{K-indicators}
	\min_{U,H}\; \|U - H\|_{F}^2, ~\st~   H \in \cH, \; U \in \cU = \left\{\hU R ~|~ R\in\R^{k\times k}, \; R^{T}R = I\right\}
	\end{equation}
	For convenience, we will refer to this model as the K-indicators model.  Other models of course can be constructed using different distance functions in \eqref{general K-indicators}.
	
	Theorem \ref{Theorem1} reveals the relations between the K-means model (\ref{Kmeans}) and the K-indicators model (\ref{K-indicators}). Both minimize a distance between the "data space" $\cR(\hU)$ and "indicator space" $\cR(H)$, or both maximize a norm of the matrix $\hU^TH$.  In either case, $H$ varies in a set of indicator matrices.  In terms of the singular values of $\hU^TH$, the difference between K-means model (\ref{Kmeans}) and K-indicators model (\ref{K-indicators}) lies in using $l_2$-norm or $l_1$-norm.  It is important to note that the two models are distinct, thus may give distinct solutions at optimality, but quite close as is indicated by inequalities in (\ref{equivalence}).  
	

	\section{KindAP Algorithm}
	
	We seek to design an algorithm for the K-indicators model~\eqref{K-indicators} that has the potential to overcome the aforementioned scalability bottleneck. We have seen that the objective in \eqref{Kmeans} for K-means is by itself non-convex as a function of $H$. In contrast,  the objective in \eqref{K-indicators} for K-indicators is convex in $H$, representing a squared distance between two sets, $\cU$ and $\cH$, both of which are non-convex sets.  Among the two,  $\cH$ is extremely non-convex with a combinatorial structure.  On the other hand, $\cU$ is less difficult to handle.  For one thing, the projection onto $\cU$ is unique in generic cases.
	
	Our idea is to break the difficult problem of solving model \eqref{K-indicators} into solving a sequence of sub-problems that satisfy two criteria: (i) each one is easier to solve, and (ii) iteration complexity is kept at linear in $n$.   In a balance of the two criteria, we propose a semi-convex-relaxation scheme: introducing a convex relaxation to $\cH$ but keeping $\cU$ unchanged.  This leads to an intermediate problem:
	\begin{eqnarray}  \label{Relaxation}
	\min_{{U,N}} \; \|U - N\|_{F}^2, ~\st~  U \in \cU, \;  N \in \cN
	\end{eqnarray}
	where $\cN =  \{N \in \R^{n \times k}| \;  0 \le N \le 1\}$ is a closed convex
	set whose boundary contains $\cH$.   
	
	
	The projection onto $\cN$ is trivial, while the projection onto $\cU$ is the so-called Procrustes problem.  The closed forms of the two projections are collected into Proposition \ref{thm:proj_H} below.
	\begin{Proposition} \label{thm:proj_H}
		The projection of matrix $U \in \R^{n \times k}$ onto the set $\cN$ is given by 
		\begin{equation}\label{Proj_N}
		\cP_{\cN}(U) = \max(\textbf{0},U),
		\end{equation} 
		and the projection of matrix $N \in \R^{n \times k}$ onto the set $\cU$ is given by 
		\begin{equation} \label{Proj_U}
		\cP_{\cU}(N) = U(PQ^T)
		\end{equation}
		where $U \in \cU$ is an arbitrary orthonormal basis of $\cR(\hU)$, and $U^TN = P \Sigma Q^T$ is a singular value decomposition of the matrix $U^TN \in \R^{k \times k}$.
	\end{Proposition}
	
	An alternating projection algorithm \cite{von1950functional} appears a natural choice for attacking the semi-relaxation model \eqref{Relaxation}, for which the computational complexity of the two projections, onto $\cU$ and $\cN$, remains linear with respect to $n$.  
	
\begin{figure}[htbp]
		\centering
		\includegraphics[width=.45\linewidth]{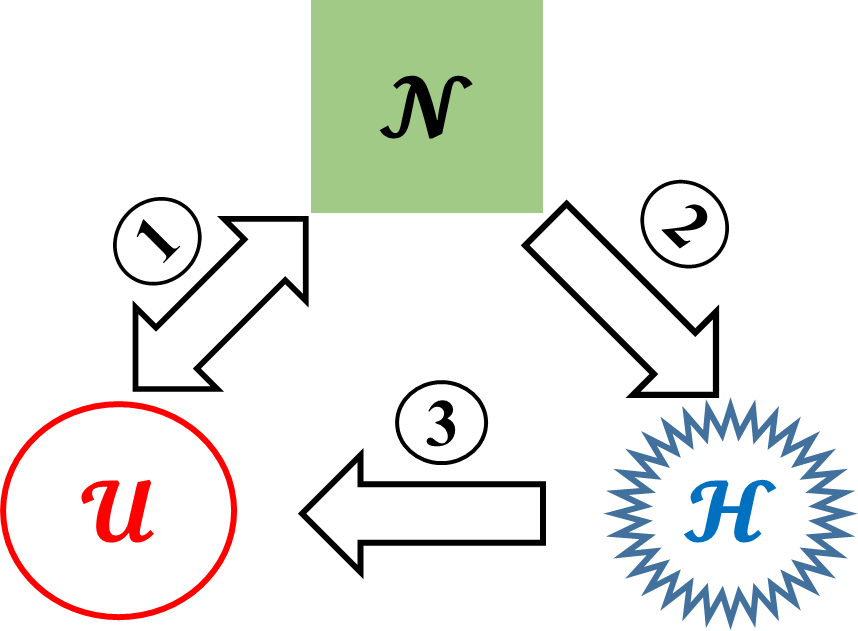}
		\caption{Big picture of KindAP: 
			Step~1 is inner alternating projection iterations for solving (\ref{Relaxation}).
			Step~2 is a rounding procedure to convert the solution of Step~1
			in $\cN$ into an indicator matrix by keeping only one nonzero, the
			largest, for each row.  Step~3 projects the indicator matrix back
			to $\cU$ to restart a new outer iteration.}
		\label{fig:big}
	\end{figure}

	On top of the above semi-convex-relaxation scheme, we construct a double-layered alternating projection framework for approximately solving the K-indicators model \eqref{K-indicators}.  See Fig.\ref{fig:big} for a schematic description.  In our algorithm, each outer iteration consists of a loop going from $\cU$ to $\cH$ and then coming back. The route from $\cU$ to $\cH$ takes a detour to $\cN$ by solving the semi-convex model (\ref{Relaxation}) via alternating projections, which are called inner iterations.   The inner or the outer iteration  is  stopped once a prescribed amount of improvement in the relevant objective value is no longer observed.  We name this algorithm KindAP (K-indicators by Alternating Projections).  
	
	\section{Related Works}
	
	This section clarifies the relationship between our study and other relevant works.  In the framework of spectral clustering, an approach called Program of Optimal Discretization (POD) is proposed to compute a binary indicator matrix   $B$ and a rotation matrix $R$ from an input matrix $\hU\in\R^{n\times k}$ consisting of $k$ leading eigenvectors of a normalized Laplacian matrix \cite{yu2003multiclass}.  The model is
	\begin{equation}  \label{SR}
	\min_{B,R} \quad  \|\hU R - B\|_F^2 	\; ~\st~   B \in \mathcal{B}, \; R^TR = I 
	\end{equation}
	where $\mathcal{B}=\{B\in\{0,1\}^{n\times k}: B\vones_k = \vones_n\}$ is the set of binary indicator matrices.
	The model aims to find a rotation matrix $R$ to best match $\hU R$ by a binary indicator matrix $B$. The POD model is also called spectral rotation (SR) in \cite{huang2013spectral}.  An ``alternating projection" type algorithm for solving the SR model was proposed.  For a fixed $R$, the binary indicator matrix $B$ is computed by
	\begin{equation} \label{NMS}
	B_{ij}=\left\{
	\begin{aligned}
	1, \quad & \mathrm{~if~} \;  j=\arg\min_{j^{\prime}}  \|u_i - r_{j^{\prime}}\|_2\\
	0, \quad & \mathrm{~otherwise~}
	\end{aligned}
	\right.
	\end{equation} 
	where $u_i$ is the $i$-th row of $U$ and $r_{j'}$ is the $j'$-th row of $R^T$.
	For a fixed $B$, the rotation matrix $R$ is given by
	\begin{equation}
	R = QP^T
	\end{equation}
	where $P$ and $Q$ are formed by the left and right singular vectors of $\hU^TB$, respectively.
	
	There are two main differences between the SR and the K-indicators clustering approaches.  The first one is about the two models which do look rather similar in appearance.  Mathematically, the K-indicators framework is based on subspace matching, that is, minimizing a distance or a measure of principle angles between two subspaces.  On the other hand, the motivation of the SR model was to add the orthonormal restriction to the matrix $R^T$ whose rows represent $k$ centers (note that in K-means model these centers are unrestricted).  Indeed, since the columns of $B$ do not form an orthonormal basis, the SR objective does not mathematically define a subspace distance and the singular values of $\hU^TB$ are not cosines of principle angles.
	
	The second difference is about the algorithms used.  For the K-indicators model, we propose a double-layered alternating projection framework based on a semi-convex-relaxation scheme, which is essentially a deterministic algorithm.  In the SR algorithm, formula (\ref{NMS}) is still greedy in nature that makes the algorithm vulnerable to the same scalability bottleneck encountered by K-means, as we will see in Section \ref{section:numerical}.   
	
	Moreover, an extra benefit of using KindAP is that the intermediate variable $N$ in (\ref{Relaxation}) produces posteriori information to evaluate the clustering quality in the absence of ground truth knowledge.  Please see more details in supplementary materials.

	\section{Numerical Experiments}
	\label{section:numerical}
	
	KindAP performs well on synthetic datasets in terms of both quality and efficiency, as is seen in Figure \ref{fig:side}. However, we need to validate it on ``real" datasets commonly used in the literature.  This section contains results from extensive numerical experiments on many real datasets.
	
	All the algorithms used in this section are implemented and run in Matlab R2018a. To be specific, the Lloyd algorithm in use is the Matlab's {\tt kmeans} function with GPU support and the K-means++ \cite{arthur2007k} initialization strategy.   In our notation, ``KM $m$'' denotes running the Lloyd algorithm with $m$ random replications in its default setting. In this section, KM 1, KM 10, KM 30, and KM 10000 will be used. KindAP is the proposed algorithm, and KindAP+L is a combination of KindAP and Lloyd in which the former is first run and the resulting centers are used to initialize the latter without further replication. We implement the SR algorithm based on \cite{huang2013spectral}. Due to SR's sensitivity to initialization, we also run it with multiple random replications and output the best solution with the lowest SR objective value.  Similarly, we use the term ``SR $m$'' to denote running the SR algorithm with $m$ random replications. Our code is available at {\tt https://github.com/yangyuchen0340/Kind}. 
	This package also contains a Python implementation, which is consistent with the well-known {\tt sklearn} package containing tools for data mining and data analysis.   We remark that the package supports SR and KindAP in the same function, and allows a flexibility in selecting different types of indicator matrices for KindAP.
	
	\subsection{Deterministic behavior of KindAP}
	
\begin{figure}[htbp]
		\begin{center}
			\subfloat[YaleB]{\includegraphics[width=0.45\textwidth]{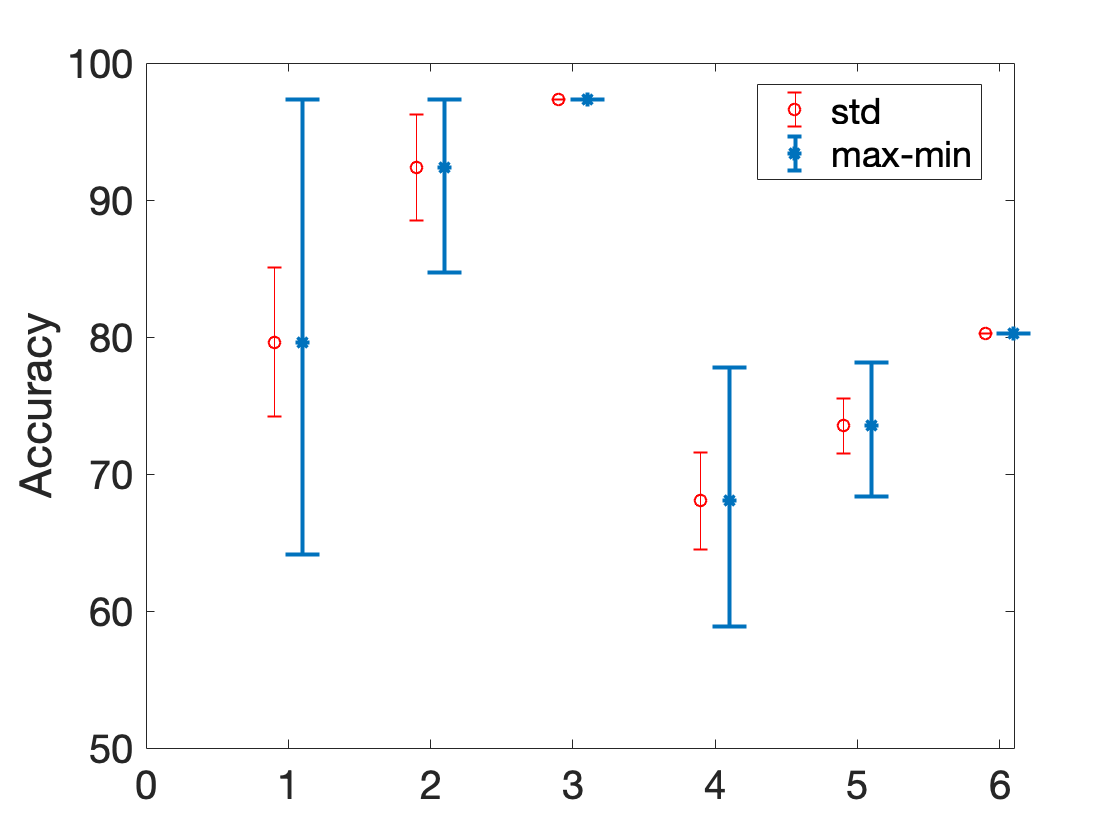}}
			\subfloat[COIL100]{\includegraphics[width=0.45\textwidth]{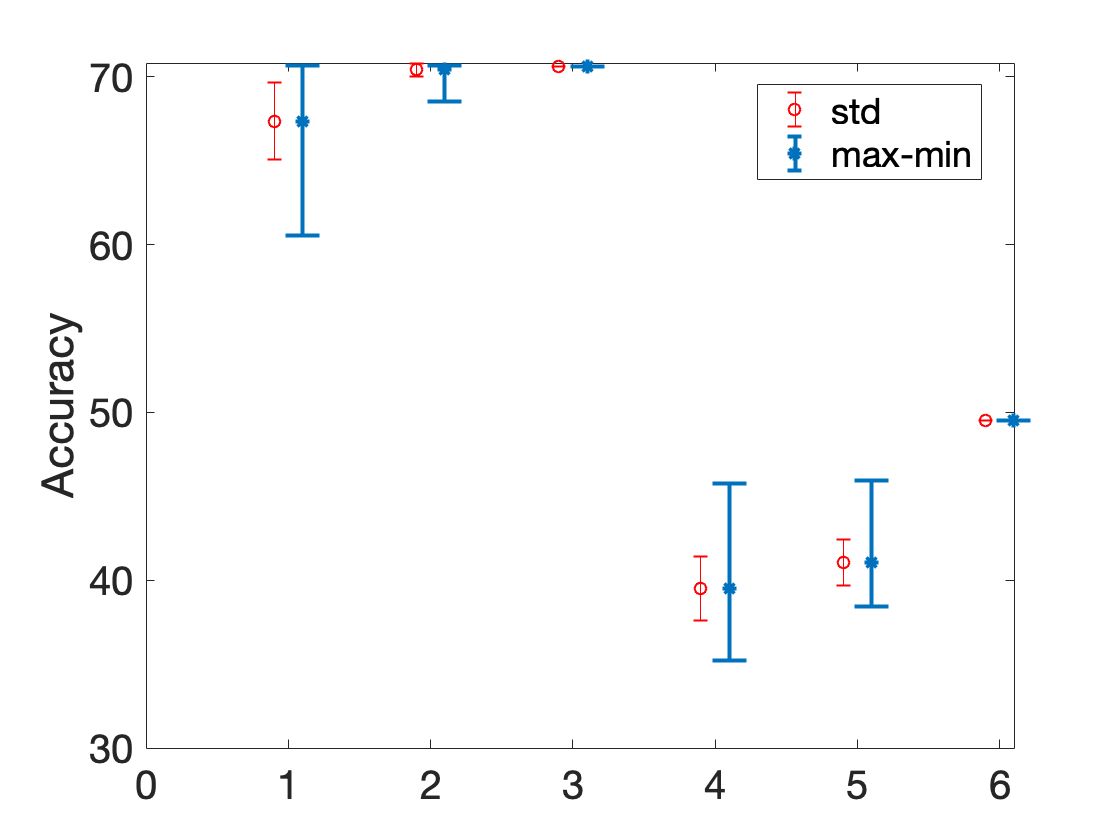}} 
			\vspace{.1cm}
			\caption{The distribution of clustering accuracy by 6 different algorithms with 200 random runs. Algorithms from left to right: SR 1, SR 10, KindAP, KM 1, KM 10, KindAP+L}
			\label{fig:DSC}
		\end{center}
	\end{figure}

	
	First, we show that KindAP algorithm is robust and essentially deterministic, as is demonstrated on two datasets YaleB and COIL100, both preprocessed by a technique called Deep Subspace Clustering \cite{ji2017deep}.  Six algorithms are tested, each with 200 random runs.  The maximum, minimum, and the standard deviation of clustering accuracy are plotted in Figure \ref{fig:DSC}.
	
	As is expected, Figure \ref{fig:DSC} shows that random replications can reduce the deviation of accuracy for both K-means and SR.   Most importantly, we observe that KindAP and KindAP+L are essentially deterministic on these two examples, achieving the identical clustering result in 200 independent random runs.  Interestingly, on both these two examples the K-indicators model happens to give much better clustering accuracy than the K-means model does.  Moreover,  when solving the K-means model for these two examples, KindAP+L produces higher accuracy than both KM~1 and KM~10 with 200 random replications (2000 replications in total for the latter).

	\subsection{Results on 35 real datasets}
	In this set of experiments, we compare the clustering accuracy and the K-means objective values between KindAP+L and KM 10000 on 35 real datasets. They include many UCI datasets, image and NLP datasets, with $k$ ranging from $k=2$ to $k=120$.   Although the underlying structures of these datasets vary from set to set, for practical reasons we preprocess all the 35 raw datasets uniformly by the normalized cut \cite{shi2000normalized, ng2002spectral} of their k-nearest-neighbor similarity graphs \cite{cai2005document}.   As a result, the clustering quality is not uniformly high.
	\begin{table}[htbp]
		\centering
		\begin{tabular}{|c|c|r|c||r|c|}
			\hline
			\multirow{2}{*}{Datasets}  & \multirow{2}{*}{$k$} &\multicolumn{2}{c||}{Clustering Accuracy }  &\multicolumn{2}{c|}{ Objective Value}   \\ \cline{3-6} \cline{3-6}
			&  &   KindAP+L   &  KM 10000   &   KindAP+L  &  KM 10000   \\ \hline \hline
			
			YaleB      &  38   &   \textbf{36.54\%}     &     36.50\%  
			&   6.0864e+00  &  \textbf{6.0858e+00} \\ \hline
			ORL        &  40   &   \textbf{67.00\%}     &     66.50\%   
			&   6.3029e+00  &  \textbf{6.0932e+00} \\ \hline
			Reuters    &  65   &   \textbf{39.36\%}     &     39.26\%
			&	\textbf{9.4345e+00}  &  1.0090e+01 \\ \hline
			PIE        &  68   &   16.17\%     &     \textbf{16.90\%}
			&   \textbf{1.0711e+01}  &  1.2179e+01 \\ \hline
			FERET      &  72   &   \textbf{66.20\%}     &     66.20\%
			&   \textbf{1.3391e+01}  &  1.3398e+01 \\ \hline
			AR          & 120   &   \textbf{60.48\%}     &    58.33\%
			&   \textbf{1.5452e+01}  &  1.7324e+01 \\ \hline
		\end{tabular}
		\smallskip
		\caption{Comparison of Clustering performance on 6 datasets with $k \ge 38$. Results with the highest clustering accuracy or the lowest objective value are highlighted in bold.}
		\label{tab:6real}
	\end{table}
	
	It turns out that on the 29 datasets with $k \le 20$, KindAP+L and KM 10000 have obtained the identical clustering accuracies, presumably corresponding to the global optima for the given datasets (see more details in the supplementary material). 
	
	The results for the remaining six datasets with $k \ge 38$ are reported in Table \ref{tab:6real}.   We observe that on the two datasets with $k\le 40$, KM 10000 reached smaller K-means objective values, but on the four datasets with $k\ge 65$ KindAP obtained better objective values, in fact significantly better in 3 out of the 4 cases.    The fact that a single KindAP run provides better initializations than 10000 random replications by the state-of-the-art K-means++ scheme \cite{arthur2007k} speaks volume for the merit of the proposed semi-convex relaxation scheme used by KindAP.  Since the running time of KindAP+L is at the same order of that of a single run of the Matlab {\tt kmeans} function, in essence KindAP+L is thousands of times faster than K-means in reaching high accuracies on large-K problems.
	
	In the previous section, we claim that the iteration complexity of KindAP is linear with respect to the size of datasets, but the efficiency of KindAP also depends on the number of iterations required for convergence. In practice, we observe that KindAP only takes several outer iterations and dozens of inner iterations on the 35 datasets (see more details in the supplementary material).

	\subsection{Datasets with deep neural network features}
	
	Deep neural network (DNN) is the trend of data mining. Recently, many deep clustering techniques have been developed to achieve higher clustering accuracy on big datasets. In this section, we select 5 image datasets, ORL, CIFAR100 (train and test), COIL100 and UKBench and process them using a DNN procedure as follows.
	We input the raw data into a pre-trained neural network~\cite{guerin2018improving}, then extract features  represented by neurons at a fully-connected layer (usually second from the last).  Afterwards, we do ordinary spectral embeddings and apply KM, SR and KindAP on these DNN features. 
	
	To be specific, we extract features from Layer {\tt avg pool} of DNN {\tt Xception} to cluster ORL, COIL100 and UKBench, and from Layer {\tt fc2} of DNN {\tt VGG-16} to cluster CIFAR100, see~\cite{guerin2018improving}, with weights pre-trained on {\tt ImageNet}. All network architecture and pre-trained weights are provided by the Python Deep Learning library {\tt Keras}.  Since data augmentation techniques and batch sizes do not make much differences to the final results, we select the default settings. Lastly, we do 30 random replications for K-means and SR. 
	
	\begin{table}[htbp] 
		\centering
		\begin{tabular}{|c||c|c|c||c|c|}
			\hline
			Datasets  & $k$ & KindAP & SR 30  & KindAP+L & KM 30 \\ \hline\hline
			ORL         & 40  & 86.50\%/{\bf 0.04} & 86.50\%/0.21 &86.25\%/0.05& 86.25\%/0.57\\ \hline
			CIFAR100(train) &  100  &   {\bf 99.63\%}/{\bf 3.75} &  84.80\%/16.51 &  99.62\%/6.53 & 94.40\%/113.22  \\ \hline
			CIFAR100(test)  & 100   &   {\bf 68.98\%}/{\bf 0.72} &  67.85\%/4.45  &  65.48\%/1.59 & 61.54\%/ 52.23 \\ \hline
			COIL100  & 100 & {\bf 98.71\%}/{\bf 1.15}   &  81.14\%/ 3.61 &  98.70\%/1.53  &  97.67\%/ 16.58  \\ \hline
			UKBench & 2550 &89.67\%/{\bf 4034} & 82.40\%/ 4727 & {\bf 89.93\%}/4602 &84.62\%/17270\\ \hline
		\end{tabular}
		\smallskip
		\caption{Clustering accuracy and timing based on pre-trained DNN features}
		\label{tab:DNN_acc}
	\end{table}
	Table \ref{tab:DNN_acc} summarizes the performance of the four algorithms on clustering the DNN features. It shows that KindAP and KindAP+L are generally more accurate than SR and K-means with 30 random replications.  
	We reiterate that our study is not about preprocessing techniques but about clustering methodologies.  However, the use of DNN techniques that produce clustering-friendly features does enable us to better evaluate the performance of different clustering methods.  On poorly processed data, one would hardly be able to differentiate behaviors of methods because all of them would produce almost equally poor clustering results.
	
	In terms of timing, KindAP is usually slower than the average running time per replication of either Lloyd or SR, but at the same order. 
	However,  both Lloyd and SR require multiple replications in order to have a chance to reach an accuracy level comparable with that of KindAP (sometimes they could only reached a lower level of accuracy after a huge number of replications).  As is indicated by the results in Table \ref{tab:DNN_acc}, KindAP runs much faster than KM 30 and SR 30 while attaining higher accuracies.  In fact, the current version of the KindAP algorithm is still far from optimal in efficiency, and we are working on new algorithms to accelerate the solution time in solving the semi-convex relaxation model. 
	
	To summarize our numerical experiments, we list several observations.  (i) KindAP and KindAP+L are essentially deterministic without the need for random replications. (ii) On small-K problems, KindAP+L works as well as the classic K-means with very large numbers of replications.  (iii) On large-K problems, KindAP and KindAP+L generally outperform their counterparts SR and Lloyd with multiple replications.   (iv) The advantages of KindAP appears more pronounced with high dimensional but clustering-friendly features extracted by advanced DNN techniques.	
	
	\section{Conclusions}

	Data clustering usually consists of two tasks: first extracting suitable features and then applying a clustering method.  The focus of this work is on the latter task for which the method of choice has arguably been K-means.  We propose the K-indicators framework \eqref{general K-indicators} that includes the classic K-means model \eqref{Kmeans} and the particular K-indicators model \eqref{K-indicators} corresponding to two different subspace distances. We promote the K-indicators model \eqref{K-indicators} because it has a convex objective function and allows an effective semi-convex-relaxation scheme, leading to the construction of an efficient algorithm called KindAP, which is essentially deterministic without any need for random replications.   Like the K-means algorithm, KindAP keeps the per-iteration complexity linear in terms of the sizes of datasets, making it practical for big-volume data clustering.

	For synthetic data with separable clusters, experiments show that K-indicators can overcome the big-K bottleneck suffered by K-means (see Fig.~\ref{fig:side}).  Is this advantage really relevant in real-world applications?  Our experiment results in Tables~\ref{tab:6real}-\ref{tab:DNN_acc} strongly suggest an affirmative answer. On the one hand, more and more big data applications come with large K values.  On the other hand, the advances in feature extraction techniques, especially those using deep neural networks, make it possible to generate clustering-friendly or even nearly separable clusters in feature spaces.  Therefore, a deterministic clustering method like KindAP that is scalable to K and linear in the the dataset size will clearly become more desirable than K-means that suffers from the big-K bottleneck due to its over-sensitivity to initializations.  
	
	We note that the K-means model \eqref{Kmeans} and the K-indicators model \eqref{K-indicators} are two distinct models that in general produce different clustering results at optimality. However, the two models are close enough (see \eqref{equivalence}) so that KindAP results can be used to initialize K-means.  Our experiments show that a single KindAP initialization can generate better clustering results, as measured by the K-means objective, than those generated from large numbers of replications using the current state-of-the-art  initialization.  A limitation of KindAP is that it requires, at least in theory, the input data matrices to be orthogonal, which is always the case in spectral clustering or similar settings. 
	
	Finally, we mention that the development of a theoretic foundation for the KindAP algorithm is an on-going effort that will be reported in a future work.
	
	\subsubsection*{Acknowledgments}
	
	Y. Yang and Y. Zhang would like to acknowledge the support from NFS Grant DMS-1418724 and from the Shenzhen Research Institute of Big Data (SRIBD). The work of L. Xu is partially supported by a Key Project of the Major Research Plan of NSFC (Grant No. 91630205)
	


	\newpage	
	
	\section{Supplementary Material}
	\subsection{Proof of Theorem 3.1}
	
	\begin{proof}
		It is easy to verify that $\dist(\cR(\hU),\cR(H)) := \min_{R^TR = I} \{\|\hU R - H\|_{F} : R \in \R^{k \times k}\}$ satisfies the non-negativity, identity of indiscernibles and symmetricity, and it suffices to show $\dist$ satisfies triangle inequality.
		
		First, let $V^{i} \in \R^{n \times k}$ for $i = 1,...,3$ are three orthnormal matrices, and denote $R^{ij}$ as the optimal solutions of the following optimization problems:
		\begin{equation*}
		R^{ij} = \argmin_{R^{T}R =  I} \|V^{i}R - V^{j}\|_{F}  \quad \; \mathrm{for} \quad 1 \le i,j \le 3, \quad \mathrm{and} \quad i \neq j.
		\end{equation*}
		Then, we have
		\begin{eqnarray*}	
			\dist(\cR(V^{1}),\cR(V^3)) 	&=&  \|V^1R^{13} - V^3\|_{F} \\
			&\le& \|V^1R^{12}R^{23} - V^3\|_{F}\\  	
			&=&  \|V^1R^{12}R^{23} - V^2R^{23} + V^2R^{23}- V^3\|_{F} \\
			&\le&  \| V^1R^{12} - V^2\|_{F}  +\| V^2R^{23}- V^3\|_{F} \\
			&=&  \dist(\cR(V^1),\cR(V^2)) + \dist(\cR(V^2),\cR(V^3)) 
		\end{eqnarray*}
		
		Therefore, the function $\dist(\cR(\hU),\cR(H)) := \min_{R^TR = I} \{\|\hU R - H\|_{F} : R \in \R^{k \times k}\}$ defines a distance between $\cR(\hU)$ and $\cR(H)$.
		
		For fixed $H\in\cH$, the closed form solution of the optimization problem $\min_{R^{T}R =  I} \|\hU R - H\|_{F}$ is given by $R^*= PQ^T$, where $\hU^TH = P \Sigma Q^T$ is a singular value decomposition of $\hU^TH$, and its optimum value is given by 
		\begin{eqnarray*} 
			\min_{R^{T}R =  I} \|\hU R - H\|^2_{F} &=&  \|\hU R^*\|_{F}^2 +  \|H\|_{F}^2 - 2 \Tr({R^*}^T\hU^TH)\\
			&=& 2k - 2\Tr[(PQ^T)^T P\Sigma Q^T)]\\
			&=& 2k -2\Tr (\Sigma) \\
			&=& 2k -2\|\hU^TH\|_* \\
			&=& 2k -2\sum_{j=1}^k\sigma_j(\hU^TH)
		\end{eqnarray*}
		Therefore,
		\begin{eqnarray}
		\min_{H \in \cH} \; \dist^2(\cR(\hU),\cR(H))&=&\min_{H\in\cH}\min_{R^TR=I} \|\hU R - H\|^2_{F}\\ &\Leftrightarrow& \max_{H\in\cH} \| \hU^TH\|_* ~~\Leftrightarrow~~ \max_{H\in\cH} \sum_{j=1}^k\sigma_j(\hU^TH).
		\end{eqnarray} 
		Moreover, 
		\begin{eqnarray*}	
			\frac{1}{2} \|\hU \hU ^T -HH^T\|_F^2
			&=& k - \|\hU^TH\|_F^2 \\  
			&=& k - 2\Tr(\hU^TH) + k \\
			&-&  (k - 2 \Tr(\hU^TH)+ \|\hU^TH\|_F^2)\\  
			&=&  \|\hU- H\|_{F}^2 -  \|I - \hU^{T}H\|_{F}^2 \\
			&\le&  \|\hU - H\|_{F}^2 
		\end{eqnarray*}
		
		Note that the above inequality always holds when we replace $\hU$ by $\hU R$ for any orthogonal $R$, therefore 
		\begin{equation}
		\frac{1}{2}\|\hU\hU^T - HH^T\|_{F}^2 \le  \min_{R^TR = I} \|\hU R - H\|_{F}^2
		\end{equation}
		
		The singular values of $\hU^TH$ are the cosines of the principle angles between $\cR(\hU)$ and $\cR(H)$, so $0 \le \sigma_j \le 1$, which implies $\sigma_j^2 \le \sigma_j$. This claim can also be derived through $\sigma_j\le||\hU^TH||_2\le||\hU||_2||H||_2=1$. Therefore, we prove the right part of inequality (\ref{equivalence}).
		\begin{equation}
		\frac{\sqrt{2}}{2}\|\hU\hU^T - HH^T\|_F \le \min_{R^TR = I} \|\hU R - H\|_{F}  \le \|\hU\hU^T - HH^T\|_F
		\end{equation}
	\end{proof}

	\subsection{Identical performance on 29 small-K datasets}
	
	Table \ref{Datasets} gives the characteristics of the 35 datasets used in the paper. Table \ref{tab:29real} compares the performance of KindAP+L and KM 10000 on 29 ``small-K'' ($k\le 20$) datasets.
	
	
	\begin{center}
		
		\begin{longtable}{|c||c|c|c|}
			
			\hline
			Dataset          & No. of Clusters   & No. of Samples  & No. of Attributes      \\ \hline\hline
			Australian       &      2            &    $  690$      &     $   14$            \\ \hline
			Breast           &      2            &    $  699$      &     $   10$            \\ \hline
			Chess            &      2            &    $ 3196$      &     $   36$            \\ \hline
			Crx              &      2            &    $  690$      &     $   15$            \\ \hline
			Diabetes         &      2            &    $  768$      &     $    8$            \\ \hline
			Heart            &      2            &    $  270$      &     $   13$            \\ \hline
			Isolet           &      2            &    $ 1560$      &     $  617$            \\ \hline
			Monk1            &      2            &    $  432$      &     $    6$            \\ \hline
			Pima             &      2            &    $  768$      &     $    8$            \\ \hline
			Vote             &      2            &    $  435$      &     $   16$            \\ \hline
			Cars             &      3            &    $  392$      &     $    8$            \\ \hline
			Iris             &      3            &    $  150$      &     $    4$            \\ \hline
			Lenses           &      3            &    $   24$      &     $    4$            \\ \hline
			Waveform-21      &      3            &    $ 2746$      &     $   21$            \\ \hline
			WINE             &      3            &    $  178$      &     $   13$            \\ \hline
			Auto             &      6            &    $  205$      &     $   25$            \\ \hline
			Control          &      6            &    $  600$      &     $   60$            \\ \hline
			Dermatology      &      6            &    $  366$      &     $   34$            \\ \hline
			glass            &      6            &    $  214$      &     $    9$            \\ \hline
			Solar            &      6            &    $  323$      &     $   12$            \\ \hline
			Segment          &      7            &    $ 2310$      &     $   19$            \\ \hline
			ZOO              &      7            &    $  101$      &     $   16$            \\ \hline
			Ecoli            &      8            &    $  336$      &     $  343$            \\ \hline
			Yeast            &     10            &    $ 1484$      &     $ 1470$            \\ \hline
			JAFFE            &     10            &    $  180$      &     $ 4096$            \\ \hline
			USPS             &     10            &    $ 9298$      &     $  256$            \\ \hline
			MNIST6000        &     10            &    $ 6000$      &     $  784$            \\ \hline
			YALE             &     15            &    $  165$      &     $ 4096$            \\ \hline
			COIL20           &     20            &    $ 1440$      &     $ 1024$            \\ \hline
			YALEB            &     38            &    $ 2414$      &     $ 1024$            \\ \hline
			ORL              &     40            &    $  400$      &     $ 4096$            \\ \hline
			Reuters          &     65            &    $ 8293$      &     $18933$            \\ \hline
			PIE              &     68            &    $11554$      &     $ 1024$            \\ \hline
			FERET            &     72            &    $  432$      &     $10304$            \\ \hline
			AR               &    120            &    $ 1680$      &     $ 2000$            \\ \hline
			\caption{The characteristics of 35 Datasets}\label{Datasets}\\
		\end{longtable}
	\end{center}
	
	\newpage
	\begin{center}
		\begin{longtable}{|c|c|r|c||r|c|}
			\hline
			\multirow{2}{*}{Datasets}  & \multirow{2}{*}{$k$} &\multicolumn{2}{c||}{Clustering Accuracy }  &\multicolumn{2}{c|}{ Objective Value}   \\ \cline{3-6} \cline{3-6}
			&  &   KindAP+L   &  KM 10000   &   KindAP+L  &  KM 10000   \\ \hline \hline
			
			Australian      &  2   &   68.70\%   &      68.70\%    
			&   1.530017e-01  & 1.530017e-01 \\ \hline
			Breast        &  2  &  51.94\%   &      51.94\%
			&    2.413292e-02  & 2.413292e-02 \\ \hline
			Crx    &  2   &   53.33\%   &      53.33\%
			& 1.304456e-01  & 1.304456e-01 \\ \hline
			Diabetes        &  2   &   64.71\%   &      64.71\%
			&   6.455858e-02  & 6.455858e-02 \\ \hline
			Heart      &  2   &   62.22\%   &      62.22\%
			&    3.106905e-01  & 3.106905e-01 \\ \hline
			Isolet          & 2   &   59.29\%   &      59.29\% 
			&   1.568688e-01  & 1.568688e-01 \\ \hline
			Monk1      &  2   &   66.90\%   &      66.90\%
			&  1.568688e-01  & 1.568688e-01 \\ \hline
			Pima        &  2   &   65.36\%   &      65.36\%
			&   3.827793e-01  & 3.827793e-01 \\ \hline
			Vote    &  2   &   56.78\%   &      56.78\%
			&	1.318960e-01  & 1.318960e-01 \\ \hline
			Cars        &  3   &   67.60\%   &      67.60\%
			&   2.180936e-01  & 2.180936e-01 \\ \hline
			Iris     &  3   &   67.60\%   &      67.60\%
			&    2.244802e-01  & 2.244802e-01 \\ \hline
			Lenses         & 3   &   41.67\%  &      41.67\%   
			&    6.056203e-01  & 6.056203e-01 \\ \hline
			Waveform-21      &  3  &   52.37\%   &      52.37\%
			&  3.746717e-01  & 3.746717e-01 \\ \hline
			WINE        &  3   &   61.80\%   &     61.80\%
			&   3.761429e-01  & 3.761429e-01 \\ \hline
			Auto    &  6   &   32.68\%   &      32.68\%
			&	5.431086e-01  & 5.431086e-01 \\ \hline
			Control        &  6   &   58.33\%   &      58.33\%
			&   4.644075e-01  & 4.644075e-01 \\ \hline
			Dermatology      &  6   &   95.90\%   &      95.90\%
			&   3.648208e-01  & 3.648208e-01 \\ \hline
			glass          & 6   &   50.47\%   &      50.47\% 
			&   1.135033e+00  & 1.135033e+00  \\ \hline
			Solar      &  6   &   37.15\%   &      37.15\%
			&   5.693965e-01  & 5.693965e-01  \\ \hline
			Segment        &  7   &   41.43\%   &      41.43\% 
			&   5.369213e-01  & 5.369213e-01  \\ \hline
			ZOO    &  7   &   51.49\%   &      51.49\%
			&	4.726120e-01  & 4.726120e-01 \\ \hline
			Ecoli        &  8   &   56.55\%   &      56.55\%
			&   2.058497e+00  & 2.058497e+00 \\ \hline
			Yeast      &  10   &   31.74\%   &      31.74\%
			&   8.848088e-02  & 8.848088e-02 \\ \hline
			JAFFE          & 10   &   62.22\%   &      62.22\%
			&   1.351849e+00  & 1.351849e+00 \\ \hline
			USPS        &  10   &   66.78\%   &      66.78\% 
			&   1.394255e+00  & 1.394255e+00\\ \hline
			MNIST6000    &  10   &    63.38\%   &      63.38\%
			&	1.777071e+00  & 1.777071e+00 \\ \hline
			YALE        &  15   &   56.36\%   &      56.36\%
			&   1.855949e+00  & 1.855949e+00 \\ \hline
			COIL20      &  20   &   82.01\%   &      82.01\%
			&   1.365658e+00  & 1.365658e+00  \\ \hline
			\caption{Identical clustering performance of two algorithms on 29 small-K datasets. }\label{tab:29real} \\
		\end{longtable}
		
	\end{center}

	Table \ref{Convergence behavior} demonstrates the convergence behaviors of KindAP on the 35 datasets. For each outer iteration (from $\cU$ to $\cH$ and back to $\cU$), we record the number of inner iterations required for convergence (alternating projections between $\cU$ and $\cN$), and the total numbers of outer and inner iterations (in parentheses) are reported in the last column. 
	
	We observe that in KindAP the first outer iteration takes the most number of inner iterations.  In fact, the first outer iteration also makes the most significant progress.  In particular, on the datasets Breast and Chess KindAP converges after only one outer iteration in which the K-indicators objective arrives at the global optimal value zero (subject to round-off error).  
	
	We also did another experiment in which after each KindAP outer iteration, we calculate cluster centers from the current solution and start the Lloyd algorithm to solve the K-means model.  Interestingly, on all the 29 datasets in Table \ref{tab:29real} the Lloyd algorithm returns identical K-means results after the first KindAP outer iteration.   That is, for solving the K-means model, only one KindAP outer iteration is necessary to generate a set of centers that allows the Lloyd algorithm to produce the presumably global optimum without any further replication.    On the other hand, after the first outer iteration until it stops, KindAP continues to improve the K-indicators objective value.  This again highlights the fact that these two models are distinct. 
			
	\newpage
	
	\begin{center}
		\begin{longtable}{|c||c|c|c|c|c|c|c|c|}
			
			\hline
			Outer Iterations          & 1 & 2 & 3 & 4 & 5 & 6 & 7 & Total\\ \hline\hline
			Australian       & 9 & 4  & $-$ & $-$ & $-$ & $-$ & $-$& 2(13)
			\\ \hline
			Breast           & 24     & $-$ & $-$ & $-$ & $-$ &  $-$ & $-$ & 1(24) \\ \hline
			Chess           & 23 & $-$ & $-$ & $-$ & $-$ &  $-$ & $-$ & 1(23) \\ \hline
			Crx             & 17 & 3 & $-$ & $-$ & $-$ & $-$  & $-$ & 2(20) \\ \hline
			Diabetes        & 12 & 3 & $-$ & $-$ & $-$ & $-$   & $-$& 2(15) \\ \hline
			Heart            & 8 & 2 & $-$ & $-$ & $-$ & $-$  & $-$& 2(10) \\ \hline
			Isolet          & 8 & 2 & $-$ & $-$ & $-$ & $-$  & $-$& 2(10) \\ \hline
			Monk1          & 8 & 2 & $-$ & $-$ & $-$ & $-$& $-$ &  2(10) \\ \hline
			Pima            & 12 & 4 & $-$ & $-$ & $-$ & $-$ & $-$ & 2(16) \\ \hline
			Vote           & 9 & 4 & $-$ & $-$ & $-$ & $-$ & $-$&  2(13) \\ \hline
			Cars           & 8 & 2 & $-$ & $-$ & $-$ & $-$& $-$ &  2(10) \\ \hline
			Iris            & 11 & 2 & $-$ & $-$ & $-$ & $-$ & $-$& 2(13) \\ \hline
			Lenses         & 40 & 4 & $-$ & $-$ & $-$ & $-$  & $-$& 2(44) \\ \hline
			Waveform-21     & 9 & 3 & 3 & $-$ & $-$ & $-$ & $-$&  2(15) \\ \hline
			WINE            & 15 & 3 & $-$ & $-$ & $-$ & $-$ & $-$ & 2(18) \\ \hline
			Auto             & 14 & 4 & $-$ & $-$ & $-$ &  $-$ & $-$ & 2(18) \\ \hline
			Control          & 13 & 4 & $-$ & $-$ & $-$ &  $-$ & $-$ & 2(17) \\ \hline
			Dermatology       & 12 & 4 & $-$ & $-$ & $-$ &  $-$ & $-$ & 2(16) \\ \hline
			glass            & 34 & 5 & $-$ & $-$ & $-$ &  $-$ & $-$ & 2(39) \\ \hline
			Solar           & 9 & 4 & $-$ & $-$ & $-$ & $-$ & $-$ & 2(13) \\ \hline
			Segment          & 12 & 5 & $-$ & $-$ & $-$ &  $-$ & $-$ & 2(17) \\ \hline
			ZOO              & 12 & 4 & $-$ & $-$ & $-$ &  $-$ & $-$ & 2(16) \\ \hline
			Ecoli            & 12 & 4 & $-$ & $-$ & $-$ &  $-$ & $-$ & 2(16) \\ \hline
			Yeast            & 15 & 4 & $-$ & $-$ & $-$ &  $-$ & $-$ & 2(19) \\ \hline
			JAFFE            & 14 & 4 & $-$ & $-$ & $-$ &  $-$ & $-$ & 2(18) \\ \hline
			USPS             & 14 & 4 & 4 & 4 & $-$ & $-$ & $-$ & 2(15) \\ \hline
			MNIST6000        & 20 & 4 & 4 & 4 & $-$ & $-$ & $-$ &  4(32) \\ \hline
			YALE             & 25 & 4 & $-$ & $-$ & $-$ &  $-$ & $-$ & 2(29) \\ \hline
			COIL20           & 14 & 4 & $-$ & $-$ & $-$ &  $-$ & $-$ & 2(18) \\ \hline
			YALEB             & 26 & 4 & 5 & 5 & $-$ & $-$ & $-$ &  4(40) \\ \hline
			ORL              & 50 & 6 & 5 & 4 & 4 & $-$ & $-$ &  5(69) \\ \hline
			Reuters           & 37 & 4 & 4 & 4 & 4 & 4 & $-$ &  6(57) \\ \hline
			PIE               & 35 & 5 & 5 & 5 & $-$ & $-$ & $-$ &  4(50) \\ \hline
			FERET            & 48 & 5 & 5 & 5 & 5 & 5 & $-$ &  6(73) \\ \hline
			AR                & 41 & 5 & 5 & 5 & 5 & $-$ & $-$ &  5(61) \\ \hline
			\caption{The convergence behavior of KindAP on 35 Datasets}\label{Convergence behavior}\\
		\end{longtable}
	\end{center}
	
	\subsection{Uncertainty information}
	
\begin{figure}[htbp]
		\centering
		\includegraphics[width=\linewidth]{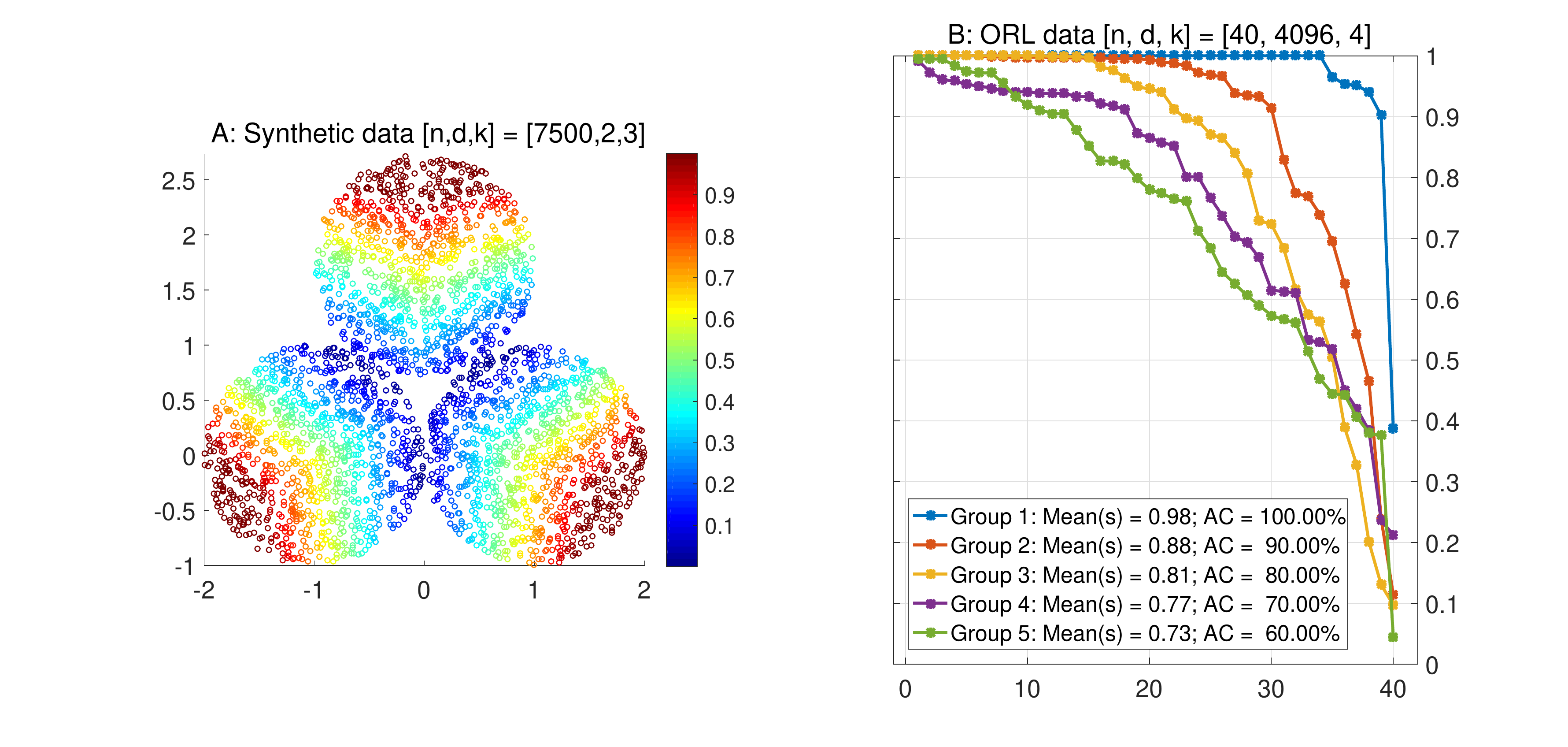}
		\caption{Soft indicators.  In plot (A), the dataset consists of three circular clusters
			in $\R^{2}$, mutually tangential to each other. Then 2500 points are uniformly
			placed inside each circle. Plot (A) shows the distribution of the points colored
			according to their KindAP-generated soft indicator values, which clearly
			are highly correlated to the distance to neighboring clusters.
			Plot (B) was generated as follows. Five groups were selected from ORL
			dataset \cite{ORL}, each containing 40 face images taken from 4 individuals
			with a varying degree of similarity.   For each group, KindAP was applied
			to the first $4$ leading eigenvectors of the normalized Laplacian of a
			similarity graph, where the similarity graph is built according to \cite{cai2005document}. Plot (B) gives the resulting
			5 soft indicators sorted in an ascending order.  Corresponding clustering
			accuracy and the mean of the soft indicator are also recorded for the 5
			groups, showing a clear correlation between the two quantities.}
		\label{fig:soft}
	\end{figure}

	In unsupervised learning, it is usually difficult to evaluate the performance due to the absence of ground truth.  People mostly use clustering accuracy and normalized mutual information, but both require ground truth. However, the proposed KindAP algorithm is able to provide some {\em a posteriori} information as a metrics of performance. 
	An output of KindAP algorithm is a non-negative matrix $N$, which generally contains more than one positive elements on each row.   It is observed in practice that the magnitude of $[N]_{ij} \ge 0$ is positively proportional to the probability that data point $i$ is in cluster $j$.  Let $\hat{N}$ hold the elements of $N$ sorted row-wise in a descending order, we define a {\em soft indicator vector} $s$, which takes values between zero and one as follows,
	\begin{equation}\label{soft-indicator}
	s_i = 1 - [\hat{N}]_{i2}/[\hat{N}]_{i1} \in [0,1], \;\;  i = 1,...,n,
	\end{equation}
	where the ratio is between the second largest and the largest elements on the $i$-th row of $N$.
	It is intuitive to expect that the closer to zero $s_{i}$ is, the more
	uncertain about the assignment to this data object, which means it is more probable that the data object $i$ belongs to more than one cluster.
	In contrast, $s_{i}$ close to one implies a safe clustering
	assignment.  This intuition is clearly validated in the simple example
	presented in Fig.~\ref{fig:soft}A
	The results in Fig.~\ref{fig:soft}B shows the relationship between
	the soft indicator and clustering accuracy on a face image dataset, ORL.  
	We observe that large soft indicator
	values on average indicate high KindAP clustering accuracy.
	This soft indicator offered by the KindAP algorithm helps address
	the very challenging issue of assessing the quality and uncertainty
	of clustering assignments in the absence of ground truth.

\end{document}